# Diffusion Transformer Meets Random Masks: An Advanced PET Reconstruction Framework

Bin Huang, Binzhong He, Yanhan Chen, Zhili Liu, Xinyue Wang, Binxuan Li,
Qiegen Liu, *Senior Member, IEEE*

*Abstract*—Deep learning has significantly advanced PET image reconstruction, achieving remarkable improvements in image quality through direct training on sinogram or image data. Traditional methods often utilize masks for inpainting tasks, but their incorporation into PET reconstruction frameworks introduces transformative potential. In this study, we propose an advanced PET reconstruction framework called Diffusion tRansformer mEets rAndom Masks (DREAM). To the best of our knowledge, this is the first work to integrate mask mechanisms into both the sinogram domain and the latent space, pioneering their role in PET reconstruction and demonstrating their ability to enhance reconstruction fidelity and efficiency. The framework employs a high-dimensional stacking approach, transforming masked data from two to three dimensions to expand the solution space and enable the model to capture richer spatial relationships. Additionally, a mask-driven latent space is designed to accelerate the diffusion process by leveraging sinogram-driven and mask-driven compact priors, which reduce computational complexity while preserving essential data characteristics. A hierarchical masking strategy is also introduced, guiding the model from focusing on fine-grained local details in the early stages to capturing broader global patterns over time. This progressive approach ensures a balance between detailed feature preservation and comprehensive context understanding. Experimental results demonstrate that DREAM not only improves the overall quality of reconstructed PET images but also preserves critical clinical details, highlighting its potential to advance PET imaging technology. By integrating compact priors and hierarchical masking, DREAM offers a promising and efficient avenue for future research and application in PET imaging. The open-source code is available at: https://github.com/yqx7150/DREAM.

This work was supported by National Natural Science Foundation of China (Grant number 62122033, 62201193), and in part by Nanchang University Interdisciplinary Innovation Fund Project (Grant number PYJX20230002). (B. Huang and B. He are co-first authors.) (Corresponding authors: B. Li and Q. Liu.)

B. Huang and B. He are with School of Mathematics and Computer Sciences, Nanchang University, Nanchang, China ({huangbin, 406100220062}@email.ncu.edu.cn).

Y. Chen is with the Ji Luan Academy, Nanchang University, Nanchang, China (6108122071@email.ncu.edu.cn).

Z. Li is with Department of Orthopedics, the First Affiliated Hospital, Jiangxi Medical College, Nanchang University, Nanchang, People's Republic of China and Jiangxi Provincial Key Laboratory of Spine and Spinal Cord Diseases, Nanchang, People's Republic of China (zhili-liu@ncu.edu.cn).

B. Li is with Institute of Artificial Intelligence, Hefei Comprehensive National Science Center, Hefei, China (libingxuan@iai.ustc.edu.cn).

X. Wang and Q. Liu are with School of Information Engineering, Nanchang University, Nanchang, China (406100240018@email.ncu.edu.cn, liuqiegen@ncu.edu.cn).

## I. INTRODUCTION

Positron emission tomography (PET) is a highly sensitive imaging technique used for diagnosing, staging, and monitoring diseases, including tumors, neurological disorders, and cardiovascular conditions [1-3]. It allows observation of metabolic processes, providing essential information for accurate medical decisions [4]. Despite its importance, PET exposes patients to ionizing radiation, posing significant health risks, particularly for vulnerable groups like pediatric patients and those requiring repeated evaluations [5-7]. Reducing radiation exposure is crucial, but lower doses often increase noise and degrade image quality, complicating clinical interpretation [8-9]. In PET images, higher noise can obscure critical diagnostic information, leading to misdiagnosis or repeat scans, undermining the goal of dose reduction [10]. Thus, advanced reconstruction methods are needed to enhance PET image quality without sacrificing diagnostic accuracy [11].

Traditional denoising methods, such as Gaussian denoising [12], total variation (TV) [13-14], and non-local means (NLM) [15-16], have been used for PET imaging. However, recent advances in deep learning, particularly convolutional neural networks (CNNs), have proven more effective. CNNs excel at detecting complex patterns in large datasets, significantly enhancing PET image quality [17-18]. For instance, Gong *et al*. [19] developed a CNN-based approach that converts PET images into full-dose equivalents, reducing noise and improving clarity. Xu *et al*. [20] proposed an encoder-decoder residual CNN architecture with skip connections, yielding satisfactory results even at low doses. Other techniques have also shown promise: Peng *et al*. [21] applied deep learning for denoising, reducing noise while preserving image details by integrating CT information, and Gong *et al*. [22] used a deep neural network for denoising PET images with simulation data and real datasets. Zhou *et al*. [23] employed a cycle-consistent GAN (Cycle GAN) to convert PET images to high-dose equivalents, improving image fidelity and diagnostic quality. Finally, Pan *et al*. [24] demonstrated the effectiveness of the diffusion model in PET image reconstruction.

In deep learning, masks are commonly used in inpainting and segmentation to guide the model's focus and improve performance. In inpainting, masks restore missing or occluded regions based on surrounding context, enabling the model to infer the missing information. For example, Iizuka *et al*. [25] used masks to facilitate image completion, emphasizing global and local consistency. Similarly, Pathak *et al*. [26] introduced the Context Encoder network, using masks to obscure parts of the image, allowing the model to learn feature representations through inpainting. Xiang *et al*. [27] reviewed deep learning methods for inpainting, highlighting the importance of dynamic masks in improving lesion restoration, especially in medical imaging, where occlusions are common. In segmentation, masks delineate regions of interest such as organs, lesions, or tumors. By focusing the model on relevant areas, masks guide the segmentation process, enabling precise identification of abnormalities. For instance, Long *et al*. [28] proposed fully convolutional networks (FCNs) using segmentation masks to delineate complex structures in biomedical images. Wang *et al*. [29] provided a survey on deep learning-based medical image segmentation, emphasizing the role of masks in improving accuracy, particularly in tumor and organ segmentation. These studies underscore the importance of masks in enhancing segmentation precision, critical for accurate diagnosis and treatment planning.

In addition to masks in inpainting and segmentation, advanced models such as transformers and diffusion models are crucial for PET image reconstruction. These models, including generative and feature extraction techniques, significantly enhance image quality. Diffusion models, known for their strong pattern coverage, are effective in denoising and improving PET images by learning data distributions through diffusion simulations. Gong *et al*. [30] showed that denoising diffusion models reduce PET image noise by incorporating MR priors and PET data-consistency constraints. Similarly, Jiang *et al*. [31] proposed an unsupervised enhancement method based on latent diffusion models, leveraging latent information to improve PET images. Transformers, valued for capturing long-term dependencies and processing large-scale data, offer notable advantages in PET reconstruction due to their strong feature extraction capabilities. For example, Luo *et al*. [32] developed a 3D transformer-GAN for high-quality PET reconstruction. Hu *et al*. [33] introduced TransEM, which uses a residual Swin-transformer-based regularization method, further validating the effectiveness of transformers in PET imaging.

Although masks have widespread applications in inpainting and segmentation tasks, their integration into PET image reconstruction has been scarcely explored. A promising direction lies in effectively incorporating the mask mechanism into powerful reconstruction models such as the diffusion transformer.

In this study, we introduce a novel mask mechanism that integrates seamlessly with the diffusion transformer model to enhance PET image reconstruction performance. Masks are applied in both the sinogram domain and the latent space, serving distinct but complementary roles. In the sinogram domain, random masks are utilized to mimic computational variability, forming enriched three-dimensional sinogram data blocks that enhance the model's ability to learn complex spatial relationships across multiple dimensions. The normal latent space is designed as a key feature of the framework, aimed at accelerating the diffusion process by operating on compact priors rather than the entire data. This mask-driven latent space is constructed from sinogram-driven compact prior and mask-driven compact prior, which provide targeted guidance during reconstruction. Additionally, a hierarchical masking strategy is employed to guide the learning process progressively. The hierarchical masks initially target localized features within the sinogram data and gradually expand their focus to capture global patterns as training progresses. This enables the model to balance fine-grained detail learning with global context comprehension effectively. Through this innovative integration of compact priors and hierarchical masking, the proposed framework not only accelerates computation but also delivers high-accuracy reconstructions, marking a significant advancement in PET image reconstruction methodologies. The theoretical and practical contributions of this work are summarized as follows:

- **Mask formulation:** We present the first study to investigate the potential of mask mechanisms in both PET sinogram domain and latent space within a diffusion transformer model. Through the design of a masked sinogram data block and a mask-driven latent space, it establishes a transformative framework that leverages masks to enhance reconstruction accuracy and feature comprehensiveness.
- **Mask utilization:** By applying high-dimensional stacking to the masked data, the method shifts from a two-dimensional to a three-dimensional framework, expanding the solution space and allowing the model to access a more diverse range of reconstruction possibilities. A hierarchical mask algorithm is further proposed, enabling the model to progressively transition from focusing on fine-grained local details to capturing broader global patterns.

The rest of the manuscript is organized as follows. Relevant background on score-based diffusion model is demonstrated in Section II. Detailed procedure and algorithm of the proposed method are presented in Section III. Experimental results and specifications about the implementation and experiments are given in Section IV. At last, conclusion is drawn in Section V.

## II. PRELIMINARY

### A. Transformer and Diffusion Models

Recent advancements in machine learning have driven significant progress in image processing, with transformers and diffusion models emerging as two powerful frameworks for addressing complex tasks. Transformers, known for their ability to capture long-range dependencies, have revolutionized feature extraction in computer vision. On the other hand, diffusion models excel in modeling intricate data distributions and iterative denoising, making them particularly effective for image restoration. This section explores these two paradigms individually and highlights their integration, which has led to breakthroughs in PET image reconstruction.

The vision transformer (ViT), introduced by Dosovitskiy *et al*. [34], revolutionized computer vision (CV) by adapting the self-attention mechanism from natural language processing to image-based tasks. The transformer model includes encoders and decoders, each consisting of multiple self-attention layers and feedforward neural networks. Self-attention captures long-range dependencies, improving contextual and relational understanding in sequences. Swin-Transformer [35] reduces the high computational complexity of traditional transformers in high-resolution image processing by introducing a hierarchical structure and local window self-attention.

The self-attention mechanism, which forms the core of transformer models, computes the attention scores using the following formula:

$$Attention(Q, K, V) = Softmax \times \left(\frac{QK^T}{\sqrt{d_k}}\right)V \quad (1)$$

where $Q$ (Query), $K$ (Key), and $V$ (Value) are linear transformations of the input features, and $d_k$ is the dimensionality of the keys for scaling. This mechanism enables the model to focus on the most relevant features in the input, making it highly effective for tasks like image restoration.

For multi-head attention, a commonly used variation in transformer models, multiple attention heads are computed and concatenated as follows:

$$MultiHead(Q, K, V) = Concat(head_1, ..., head_n)W^O \quad (2)$$

where each $head_i$ represents an independent self-attention mechanism. This approach allows the model to capture diverse features across multiple subspaces.

Transformers excel in feature extraction for image reconstruction tasks. Hu *et al.* [33] introduced TransEM, a PET image reconstruction method that incorporates a residual Swin-Transformer as a regularizer within an iterative framework. This model extracts shallow features via convolutional layers, processes them through Swin-Transformer layers, and fuses these features to enhance reconstruction quality. Cui *et al.* [36] proposed TriDo-Former, a triple-domain transformer-based model that operates across sinogram, image, and frequency domains, significantly improving PET reconstruction quality by combining denoising and spatial-frequency information.

While transformer models focus on capturing long-range dependencies and extracting high-level features, diffusion models excel in modeling complex data distributions and iterative noise removal. The integration of these two paradigms has led to significant advancements in tasks such as image synthesis and reconstruction, where each contributes unique strengths to overcome existing challenges. Diffusion models have gained prominence in image synthesis and restoration due to their ability to model complex data distributions. These models employ a forward diffusion process to gradually corrupt data into noise and a reverse process to iteratively recover data, guided by a neural network. The forward diffusion process is defined as:

$$q(x_t|x_{t-1}) = \mathcal{N}(x_t; \sqrt{1-\beta_t}x_{t-1}, \beta_t I) \quad (3)$$

where $x_t$ denotes the noised image at time-step $t$, $\beta_t$ represents the predefined scale factor, and $\mathcal{N}$ represents the Gaussian distribution. During the reverse process, diffusion models sample a Gaussian random noise map $x_t$, then progressively denoise $x_t$ until it achieves a high-quality output $x_0$:

$$p(x_{t-1}|x_t, x_0) = \mathcal{N}(x_{t-1}; \mu_t(x_t, x_0), \sigma_t^2 I) \quad (4)$$

To train a denoising network $\epsilon_\theta(x_t, t)$, given a clean image $x_0$, diffusion models randomly sample a time step $t$ and a noise $\epsilon \sim \mathcal{N}(0, I)$ to generate noisy images $x_t$ according to Eq. (4).

DDPM has demonstrated exceptional performance in generating high-quality images, making it a versatile tool for both synthesis and restoration tasks. The framework uses a forward diffusion process to corrupt data and a reverse process, guided by a trained neural network, to iteratively recover the data. The simplified training objective, focused on noise prediction, is defined as:

$$L_{simple} = E_{x_0,t,\epsilon}||\epsilon - \epsilon_\theta(x_t, t)||^2 \quad (5)$$

This objective ensures accurate noise estimation during the forward process, effectively guiding reverse diffusion for image generation. With its ability to model complex data distributions and robustness to noise variations, DDPM is particularly suited for restoration tasks. By conditioning the reverse process on degraded inputs, it adapts to tasks such as denoising and super-resolution, recovering high-fidelity outputs. Recent advancements, including noise scheduling optimization and architectural improvements, further enhance DDPM's computational efficiency, enabling its integration with transformers for challenging applications like PET image reconstruction.

Recent advances have explored the integration of transformers and diffusion models to address challenges in PET image reconstruction. By combining the feature extraction capabilities of transformers with the data modeling strengths of diffusion models, these hybrid approaches achieve superior performance in reconstruction tasks. Huang *et al.* [37] proposed a method that integrates diffusion models with transformers for generating full-dose PET images from low-dose inputs. In the first stage, the diffusion model learns an image prior representation. In the second stage, the transformer leverages these priors to directly estimate image features, resulting in stable and realistic outputs. The reconstruction loss often incorporates both data consistency and feature learning:

$$\mathcal{L} = \| y - Ax \|^2 + \lambda \| R(x) - T(x) \|^2 \quad (6)$$

where $y$ is the observed sinogram, $A$ is the system matrix, $x$ is the reconstructed image, $R(x)$ is the regularized feature from traditional methods, and $T(x)$ represents transformer-learned features. The weight $\lambda$ balances data fidelity and feature consistency.

This combination enables leveraging the long-range dependency modeling of transformers and the robustness of diffusion models, providing a structured framework for high-quality PET image reconstruction. The integration of transformers and diffusion models provides a complementary framework for high-quality PET image reconstruction. By leveraging the strengths of both paradigms, these hybrid approaches address limitations in traditional methods, achieving significant advancements in accuracy and robustness for clinical applications.

### B. Masks

Masks have played a significant role in inpainting and segmentation. These applications highlight the versatility of masks in addressing different challenges, ranging from Inpainting tasks to guiding comprehensive Segmentation tasks.

Inpainting tasks leverage masks to restore missing or occluded regions of an image based on surrounding pixel values, focusing on local details and context-aware features. Yu *et al.* [38] proposed the mask-based latent space feature inpainting (MLR) method, where masks are used to obscure portions of pixel observations. By inpainting latent features, this approach enhances the model's ability to infer complete state representations and better understand contextual relationships. Similarly, Wang *et al.* [39] introduced LocalMIM, a method that uses masks to obscure specific image regions, facilitating multi-scale feature learning during the inpainting process. Furthermore, Liu *et al.* [40] proposed an adaptive masked training framework for image inpainting tasks, where dynamic masking of input images during training improves the model's capacity to capture structural consistency and enhances its robustness across diverse occlusion patterns.

Segmentation tasks employ masks to delineate regions of interest within an image, guiding models to identify specific structures or features. Unlike inpainting, segmentation focuses on partitioning the image into meaningful regions rather than restoring missing pixels. In medical imaging, for example, masks are commonly used to label tumors or organs, enabling precise localization and characterization of abnormalities. Ronneberger *et al.* [41] introduced the U-Net architecture, which uses segmentation masks to accurately identify structures in biomedical images, proving particularly effective for tasks such as organ segmentation and tumor detection. Similarly, Long *et al.* [42] proposed fully convolutional networks (FCNs) for semantic segmentation, demonstrating how segmentation masks guide models in delineating complex boundaries and achieving pixel-level accuracy. This approach underpins many clinical applications where pixel-level precision is essential for decision-making.

To better understand the integration of masks within deep learning frameworks, it is important to define the mask operation formally. In this context, masks are represented as binary images, where each pixel value is either 0 or 1. This computationally efficient representation is suitable for guiding selective operations in reconstruction tasks. Let the original image be $I(x, y)$ and the mask be $M(x, y)$. The masked image $I'(x, y)$ is mathematically defined as:

$$I'(x, y) = I(x, y) \odot M(x, y) \quad (7)$$

In this equation, $M(x, y)$ is a binary image, where $M(x, y) = 1$ retains the pixel value from the original image, and $M(x, y) = 0$, masks the corresponding pixel. This formulation provides a foundation for integrating masks into reconstruction workflows, enabling selective feature learning and targeted data restoration.

The next section introduces our proposed method, which combines dual-random masks and diffusion-transformer synergy to establish a robust framework for high-quality PET reconstruction.

## III. PROPOSED METHOD

### A. Motivation

Masks have conventionally been used in image inpainting and segmentation tasks, where deep learning networks predict and fill masked areas to complete an image. Despite their success in these fields, their application to the complex challenges of PET image reconstruction has remained relatively unexplored. PET image reconstruction requires not only filling missing data but also preserving critical clinical details and suppressing noise in a highly sensitive imaging process. This unique complexity presents an opportunity to redefine the role of masks and expand their utility. In this study, we extend the role of masks beyond conventional usage by integrating them into both the sinogram domain and the latent space, creating a multi-level framework that enhances the reconstruction process. By incorporating mask-based strategies into advanced architectures such as the diffusion transformer model, we achieve significant improvements in noise suppression and detail preservation, providing a robust solution for high-quality PET image reconstruction.

To further strengthen the model's ability to learn and predict PET sinogram data, both random masks and progressively shrinking hierarchical masks are employed. This approach introduces a progressive learning strategy that shifts the model's focus from localized to global sinogram information. Random masks are applied to each sinogram, generating $(N − 1)$ distinct masked versions, each with a size of $16 \times 16$ pixels. The masked sinograms are then stacked with the original sinogram across channels, effectively mapping the 2D image into 3D space and allowing the model to capture high-dimensional correlations that improve predictive accuracy. This integration of mask-based strategies into PET image reconstruction establishes a novel framework that enhances model diversity, robustness, and performance across various imaging challenges.

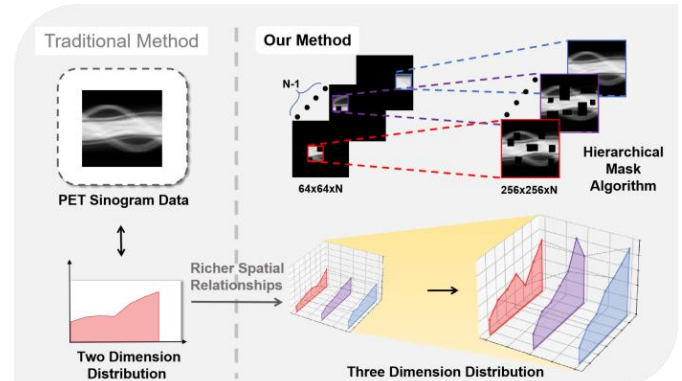

**Fig. 1.** Comparison of the traditional PET reconstruction method and the proposed DREAM framework. DREAM employs a hierarchical mask algorithm to create 3D sinogram data blocks with richer spatial relationships.

The resulting data block is then used to train the diffusion transformer model (DTM) [43]. During training, the model samples and learns from small data cubes within this block, with cube sizes progressively expanding from $64 \times 64 \times N$ to $256 \times 256 \times N$. In this process, diffusion and transformer models each contribute unique benefits. Diffusion models excel at capturing detailed local patterns, while transformer models effectively handle long-range dependencies. However, diffusion models require extensive iterative processing, leading to longer training times, whereas transformer U-Nets offer faster results but lack fine-grained detail capture. To integrate their strengths efficiently, the diffusion model's training target is shifted from full image data to compact priors in the latent space, significantly reducing training time while preserving reconstruction quality.

### B. Training Procedure

The pipeline of DREAM training procedure is illustrated in Fig. 2, detailing both data composition and model architecture. In the data composition phase, each individual sinogram undergoes random masking, where 10% of the sinogram is covered with $16 \times 16$ blocks to introduce variability. Through this process, $(N − 1)$ randomly masked sinograms are generated and then combined with the complete sinogram to form a 3D sinogram data block. During training, the hierarchical mask size gradually decreases, while the model progressively learns from larger data blocks, starting from $64 \times 64 \times N$, progressing to $128 \times 128 \times N$, $192 \times 192 \times N$, and finally $256 \times 256 \times N$. This hierarchical training strategy enhances the model's ability for comprehensive reconstruction by integrating micro-level details with macro-level features, leading to more accurate and holistic outputs.

In the model architecture phase, the training procedure incorporates the compact prior (CP) extraction block, a diffusion stage, and the transformer stage, each of which contributes to the robustness and effectiveness of the reconstruction process. Initially, to enhance the training diversity and robustness, the sinogram data undergoes a multi-stage preparation process involving random masking and high-dimensional stacking. $(N-1)$ binary random masks $M_1$ to $M_{N-1}$ of size 16×16 are applied to the sinogram data $S_0$ to generate $(N-1)$ masked sinograms:

$$S_i = S_0 \odot M_i \quad (8)$$

where, $\odot$ represents element-wise multiplication. These masked sinograms are then combined with the original sinogram to form a 3D sinogram data block:

$$S_{3D} = stack(S_1, \dots, S_{N-1}, S_0) \quad (9)$$

The resulting 3D data block $S_{3D}$ is structured as $R^{h \times w \times N}$, where $h$ and $w$ denote the height and width of the sinograms. This representation allows the model to explore complex spatial correlations across the stacked channels.

During training, hierarchical masks of gradually decreasing sizes are applied to the 3D sinogram data block, enabling the network to shift its focus from fine-grained local features to broader global patterns. The hierarchical masking process follows a predefined progression:

$$B_\vartheta = \begin{cases} 64 \times 64 \times N & if\ \vartheta \leq 100000 \\ 128 \times 128 \times N & if\ 100000 < \vartheta \leq 200000 \\ 192 \times 192 \times N & if\ 200000 < \vartheta \leq 300000 \\ 256 \times 256 \times N & if\ \vartheta > 300000 \end{cases} \quad (10)$$

where, $\vartheta$ represents the number of training iterations, and the predefined milestones correspond to every 100,000 iterations.

The noise-free and noisy sinogram data blocks are concatenated and downsampled using the PixelUnshuffle operation to serve as the input for the CP extraction network. Subsequently, sinogram compact prior (SCP) and mask-driven compact prior (MCP) are extracted by the CP extraction network and combined as $\varphi$ (SMCP) in the latent space. SMCP is then utilized as dynamic modulation parameters in the multi-head transposed attention and gated feed-forward network of the transformer stage to guide PET training:

$$A' = W_l^1 \varphi \odot \text{Norm}(A) + W_l^2 \varphi \quad (11)$$

where $\odot$ indicates element-wise multiplication, Norm denotes layer normalization, $W_l$ represents linear layer, $A$ and $A'$ are input and output feature maps respectively.

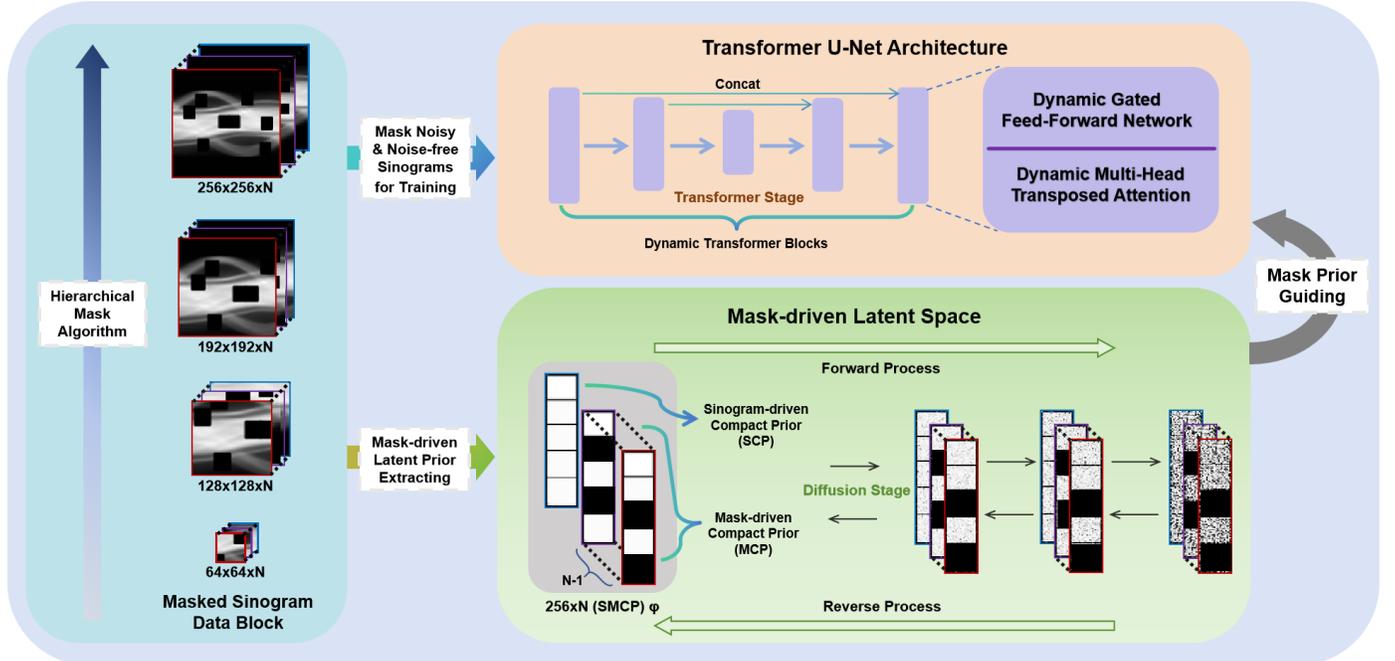

Fig. 2. The pipeline of DREAM training procedure. DREAM mainly consists by mask-driven latent space, diffusion stage and transformer stage. The random masks and hierarchical masks are combined to form the sinogram data blocks. SMCP of sinogram data blocks will be feed into the diffusion stage to predict and guide transformer stage to reconstruct final result.

In the multi-head transposed attention, global spatial information is aggregated by projecting $A'$ into query $Q = W_d^Q W_c^Q A'$, key $K = W_d^K W_c^K A'$, and value $V = W_d^V W_c^V A'$ matrices, followed by reshaping and dot-product operations to generate a transposed-attention map. This map is further processed using learnable scaling parameter $\gamma$ and channel separation to generate attention map:

$$\hat{A} = W_c \hat{V} \cdot Softmax(\hat{K} \cdot \hat{Q}/\gamma) + A \quad (12)$$

where $Softmax$ is an activation function that converts a vector of values into a probability distribution, where each value's probability is proportional to the exponential of the input value, typically used in the output layer of a classification neural network [18]. The gated feed-forward network aggregates local features by employing $1 \times 1$ Conv layers to aggregate information from different channels and $3 \times 3$ depth-wise Conv layers to aggregate information from spatially neighboring pixels. Additionally, the gating mechanism is applied to enhance information encoding. The gated feed-forward network is characterized by the following process:

$$\hat{A} = \text{GELU}(W_d^1 W_c^1 A') \odot W_d^2 W_c^2 A' + A \quad (13)$$

where GELU, Gaussian error linear unit, is an activation function that uses a smooth, non-linear transformation based on the Gaussian cumulative distribution function to activate neurons in a neural network [44]. The CP extraction block and transformer stage are jointly trained, enabling the transformer stage to effectively utilize SMCP for PET reconstruction.

In the diffusion stage, SMCP is trained using the strong data estimation ability of the diffusion stage. The CP extraction network is utilized to capture $\varphi$, which is then subjected to the diffusion process to sample $S_T$:

$$q(S_T|\varphi) = \mathcal{N}(S_T; \sqrt{\bar{\alpha}_T}\varphi, (1-\bar{\alpha}_T)x) \quad (14)$$
$$\alpha_T = 1 - \beta_t; \bar{\alpha}_T = \Pi_{i=0}^{t}\alpha_i \quad (15)$$

where $T$ is the total number of iterations, $\beta_t$ indicates the predefined scale factor.

In summary, DREAM has two training stages: the transformer stage and the diffusion stage. In the transformer stage, DREAM directly uses the SMCP of the noise-free sinogram data to guide the training of the transformer, focusing on the complete PET data. In the diffusion stage, only the SMCP is trained separately without the need to train the noise-free sinogram data.

### C. Reconstruction Procedure

DREAM initiates from the $T$-th time step and progressively conducts all denoising iterations to derive $\hat{\varphi}$, which is subsequently utilized to guide the reconstruction process. Unlike traditional diffusion models, DREAM leverages the CP extraction network and the 3D sinogram data block to enhance denoising and reconstruction efficiency. The iterative denoising process is defined as:

$$\hat{\varphi}_{t-1} = \frac{1}{\sqrt{\alpha_t}}\left(\hat{\varphi}_t - \epsilon_t \frac{1-\alpha_t}{\sqrt{1-\bar{\alpha}_t}}\right) \quad (16)$$

where symbol $\epsilon_t$ represents the noise at time step $t$, which is estimated through the CP extraction network.

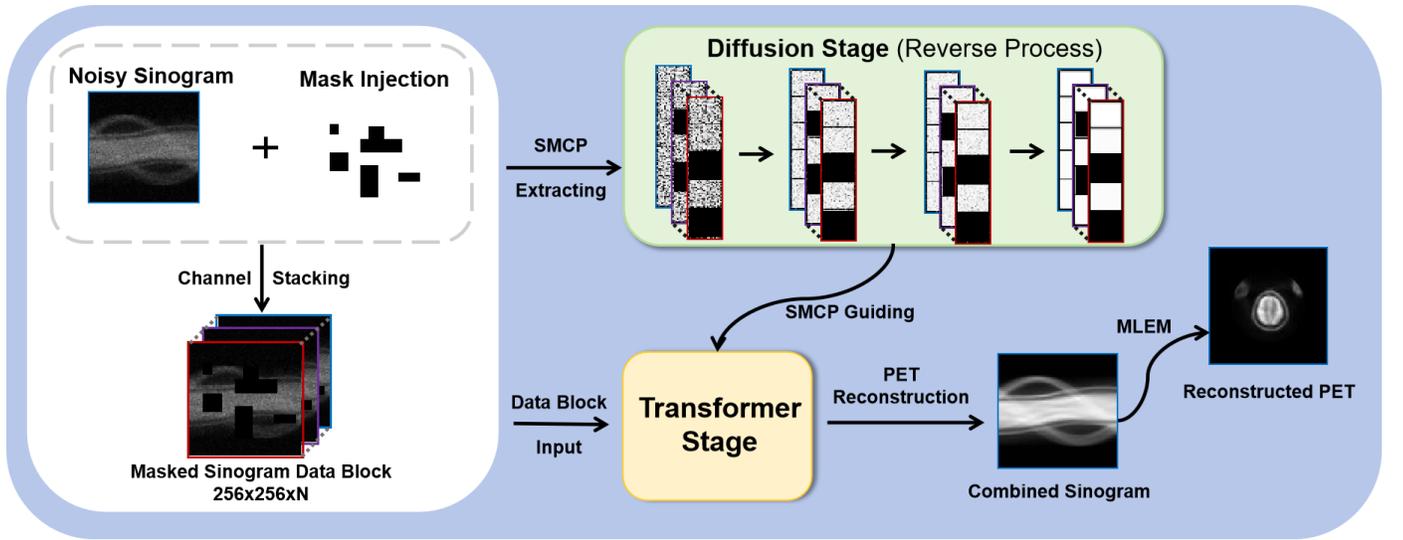

Fig. 3. The pipeline of DREAM reconstruction procedure. Injecting random masks to (N-1) noisy sinograms and stacking with a single noisy sinogram to form a noisy sinogram data block. SMCP of noise-free sinogram data block will be reconstructed through diffusion stage. Noise-free sinogram data block is reconstructed under the guidance of the SMCP and subsequently combined using weighted averaging to generate final PET sinogram. This combined PET sinogram is then processed through the MLEM algorithm to produce the final PET image.

Upon completing $T$ iterations, the final estimated SMCP $\hat{\varphi}$ is guiding to the transformer stage, which utilizes it for reconstruction optimization. The transformer stage integrates SMCP guided multi-head transposed attention and gated feed-forward networks to refine the reconstruction.

After obtaining the final reconstructed sinogram data block, it is reverted to its original form by reassembling the data block into a new 2D sinogram through weighted averaging. Finally, MLEM iterations are applied to generate data in the PET image domain.

The final output $S'$ is the reconstructed sinogram after the DREAM model's reconstruction process, which integrates both the SMCP-guided diffusion and transformer stages. The ideal PET image $I'$ is obtained using the MLEM algorithm:

$$I' = MLEM(S'), S' = DREAM(S_{all})$$
$$S_{all} = \sum_{i=0}^{N}\omega_i S_i, \sum_{i=0}^{N}\omega_i = 1 \quad (17)$$

This structured approach combines random masking and SMCP-guided reconstruction to produce high-quality PET images. $\omega_i$ are the weighting coefficients for $S_i$, satisfying the normalization constraint $\sum_{i=0}^{n}\omega_i = 1$. The term $S_{all}$ is a weighted average of the sinogram components $S_i$, where $S_0$ represents the noisy sinogram and $S_1$ to $S_N$ are masked versions generated through binary masks $M_1$ and $M_2$. This weighted formulation ensures that complementary information from both original and masked sinograms is effectively combined for enhanced reconstruction quality.

The iterative update rule for each pixel $n$ is compactly expressed as:

$$I_n^{(k+1)} = I_n^{(k)} \cdot \frac{\sum_{m=1}^{N} G_{mn} \frac{S_{all_m}}{\sum_{n=1}^{M} G_{mn} I_n^{(k)}}}{\sum_{m=1}^{N} G_{mn}} \quad (18)$$

where $I_n^{(k)}$ and $I_n^{(k+1)}$ represent the PET image intensity at the $k$-th and $(k+1)$-th iterations, respectively. $G_{mn}$ is the system matrix mapping the sinogram space to the image space. $S_{all_m}$ is the reconstructed sinogram intensity for index $m$. This formulation balances the data fidelity with the constraints imposed by the system matrix $G$. Through this SMCP guided and MLEM integrated framework, DREAM effectively reconstructs high quality PET images, providing clinically meaningful outputs without the need for hierarchical lesion-specific refinement.

Following the DREAM based reconstruction, the sinogram data block is reassembled into its original 2D form through weighted averaging. This reconstructed data is further optimized using the SMCP guided reconstruction framework, where the SMCP extracted from the 3D sinogram data block serves as a prior to enhance detail preservation and global consistency during iterative updates of the reconstruction targets. The overall optimization equation of DREAM can be expressed as:

$$S' = \underset{S}{\operatorname{argmin}}[\|I - MLEM(S_{all})\|^2 + \mu R(S_{all}, \hat{\varphi})]$$
$$s.t.\ S_{all} = \sum_{i=0}^{N} \omega_i S_i,\ \sum_{i=0}^{N} \omega_i = 1, \quad (19)$$

Here, $S'$ represents the reconstructed sinogram, serving as the optimized result of the reconstruction process. $I$ denotes the ideal image, used as a reference to evaluate the reconstruction accuracy by comparing it with $S'$. $R(S_{all}, \hat{\varphi})$ is the regularization term that incorporates SMCP information to ensure consistency and guide the reconstruction process. The operation $MLEM(S_{all})$ applies the Maximum Likelihood Expectation Maximization algorithm to reconstruct the target image from the sinogram $S_{all}$, ensuring fidelity to the observed data.

The augmented Lagrangian function for DREAM is defined as:

$$L(S', \lambda) = \|I - MLEM(S_{all})\|^2 + \mu R(S_{all}, \hat{\varphi})$$
$$+ \lambda(S_{all} - \sum_{i=0}^{N} \omega_i S_i) + \frac{\rho}{2}\|S_{all} - \sum_{i=0}^{N} \omega_i S_i\|^2 \quad (20)$$

The Lagrange multiplier $\lambda$ enforces the constraint $S_{all} = \sum_{i=0}^{N} \omega_i S_i$, ensuring that the weighted average of $S_i$ adheres to the defined relationship critical for reconstruction. To further promote consistency and minimize deviations from this constraint, the penalty parameter $\rho$ is introduced, controlling the impact of constraint violations during the optimization process. Together, these components ensure that the reconstruction maintains a balance between data fidelity and the structural relationships defined by the weights $\omega_i$.

To solve the optimization problem, we iteratively update the variables $S'^{j+1}$, and $\lambda^{j+1}$ using the following steps:

1) *Step 1: Update* $S'^{i+1}$ ← Keeping $\lambda$ fixed, the optimization problem for $S$ is given by:

$$S'^{j+1} = \underset{S}{\operatorname{argmin}}[\|I - MLEM(S_{all})\|^2 + \mu R(S_{all}, \hat{\varphi})$$
$$+ \lambda(S_{all} - \sum_{i=0}^{N} \omega_i S_i) + \frac{\rho}{2}\|S_{all} - \sum_{i=0}^{N} \omega_i S_i\|^2 \quad (21)$$

This step involves minimizing the reconstruction loss and incorporating the regularization term $R(S_{all}, \hat{\varphi})$, ensuring SMCP consistency. The updated $S'^{i+1}$ reflects both the data fidelity term $\|I - MLEM(S_{all})\|^2$ and the regularization.

2) *Step 2: Update* Lagrange Multiplier $\lambda^{i+1}$ ← The Lagrange multipliers $\lambda$ are updated to enforce the constraints:

$$\lambda^{j+1} = \lambda^j + \rho(S_{all}{}^{j+1} - \sum_{i=0}^{N} \omega_i S_i) \quad (22)$$

By iteratively performing these updates, the optimization ensures that the reconstructed sinogram $x^{i+1}$ satisfies data fidelity, SMCP consistency, and lesion-related constraints. This iterative process effectively balances global consistency and local detail preservation, ultimately leading to high-quality PET image reconstruction.

**Algorithm 1: DREAM for Reconstruction**

**Require:** $S^0, S_i^0, \lambda^0, \omega_i, \rho, \hat{\varphi}, M, \mu$
1: **Initialization:** $S'^0$, and $\lambda^0$
2: **For** $i = 0$ to convergence:
3: 　　Update $S'^{j+1}$ via Eq. (21)
4: 　　Update $\lambda^{j+1}$ via Eq. (22)
5: **End for**
6: **Final Reconstruction:** $I' = MLEM(S')$
7: **Return** $I'$

Algorithm 1 provides a detailed framework for reconstructing noisy sinogram data using the SMCP guided reconstruction process. The algorithm ensures robust optimization by iteratively updating variables while balancing data fidelity and regularization constraints to achieve high-quality PET image reconstruction.

## IV. Experiments

In this section, we introduce the implementation details of the proposed DREAM, as well as the datasets we used for evaluation. Subsequently, the reconstruction results are reported and analyzed. Both quantitative and qualitative evaluations are comprehensively conducted to investigate the performance of DREAM.

### A. Data Specification

*Patient data:* The data used in this experiment is a PET image dataset of the brain, which is acquired based on an All-Digital PET imaging system (DPET-100 platform provided by RaySolution Digital Medical Imaging Co., Ltd.) with a FOV of $500 \times 500 mm^2$ and an axial FOV of $201.6mm$. The single-layer PET image matrix is $256 \times 256$, with a pixel size of $2.0mm \times 2.0mm$, The data are all obtained from real patients' head PET/CT scans. A total of 5040 2D slices are collected from 84 brain 18F-FDG clinical patients. Among these patients, 4200 2D slices are selected for the purpose of training, while the remaining 40 2D slices from one patient are designated as a validation dataset for validation of the model during training. Apart from 84 patients, we additionally select a total of 840 2D slices of 14 patients as a test dataset to evaluate the generation performance of DREAM. These PET sclices were then forward projected to generate noise-free sinograms. To generate noisy sinograms, Poisson noise was applied. Scattered events were modeled using the SimSET package with a cylindrical scanner configuration, omitting block and gap effects. Additionally, 20% uniform random events were incorporated. For accurate quantitative imaging, attenuation maps, along with mean scatter and random distributions, were included in all reconstruction approaches. The anticipated total event count over a 1-hour duration was 8 million. Ten independent noisy realizations were simulated, with each being reconstructed separately for comparative analysis. This study is approved by the institutional review board of the Beijing Friendship Hospital, Capital Medical University, Beijing, China. The approval number is 2022-P2-314-01.

### B. Model Training and Parameter Selection

During the training phase, paired projection data were used as the input to the neural network. Specifically, random masks were applied to the first $(N-1)$ channels, while the channel $N$ remained unmodified. Additionally, a hierarchical mask was progressively reduced, enabling the model to perform incremental training from local to global information. The input image sizes were gradually increased in the following order: 64, 128, 192, and 256. The proposed method employs a four-level encoder-decoder architecture. In the transformer stage, the multi-head attention mechanism was configured with attention heads set to [1, 2, 4, 8], corresponding to channel sizes of [48, 96, 192, 384]. Furthermore, within the four levels, the number of dynamic transformer blocks was set to [3, 5, 6, 6], while the channel size for the JCP extraction blocks was fixed at 64. In the diffusion stage, the number of diffusion steps $T$ was set to 4. The Adam optimizer was used for optimization, with a learning rate of $2\times 10^{-4}$, $\beta_1 = 0.9$, and $\beta_2 = 0.99$.

### C. Quantitative Indices

To evaluate the quality of the reconstructed data, peak signal-to-noise ratio (PSNR), structural similarity index (SSIM), and mean squared error (MSE) are used for quantitative assessment.

PSNR describes the maximum possible power of the signal in relation to the noise corrupting power. Higher PSNR means better image quality. Denoting $x$ and $y$ to be the estimated reconstruction and the reference image, PSNR is expressed as:

$$PSNR(x, y) = 20\log_{10}[Max(y)/\|x-y\|_2] \quad (23)$$

SSIM is used to measure the similarity between the ground-truth and reconstruction, and it is defined as:

$$SSIM(x, y) = \frac{(2\mu_x\mu_y + c_1)(2\sigma_{xy} + c_2)}{(\mu_x^2 + \mu_y^2 + c_1)(\sigma_x^2 + \sigma_y^2 + c_2)} \quad (24)$$

where $\mu_x$ and $\sigma_x^2$ are the average and variances of $x$. $\sigma_{xy}$ is the covariance of $x$ and $y$. $c_1$ and $c_2$ are used to maintain a stable constant. MSE is employed to evaluate the errors and it is defined as:

$$MSE(x, y) = \frac{1}{W}\sum_{i=1}^{W}\|x_i - y_i\|_2^2 \quad (25)$$

where $W$ is the number of pixels within the reconstruction result. If MSE approaches to zero, the reconstructed image is closer to the reference image.

### D. Experimental Comparison

To evaluate the reconstruction performance of DREAM, we conducted a comparison with several state-of-the-art approaches. For traditional methods, we used the MLEM [45] algorithm to reconstruct the projection data. Additionally, we implemented several deep learning-based reconstruction methods using the same training dataset, including the diffusion probabilistic model-based DDPM [30], the vision transformer-based U-ViT [46], the unsupervised diffusion model NCSN++ [47], and the supervised diffusion model IR-SDE [48].

TABLE I
RECONSTRUCTION PSNR/SSIM/MSE FROM PET PROJECTION DATA USING DIFFERENT METHODS.

| Methods | MLEM | DDPM | U-ViT | NCSN++ | IR-SDE | **DREAM** |
|---|---|---|---|---|---|---|
| PSNR | 21.12 | 20.75 | 20.79 | 21.92 | 33.22 | **34.57** |
| SSIM | 0.808 | 0.787 | 0.790 | 0.839 | 0.950 | **0.955** |
| MSE | 0.0082 | 0.0090 | 0.0089 | 0.0067 | 0.0005 | **0.0003** |

The reconstruction performance was objectively quantified using several metrics, including PSNR, SSIM, and MSE. To evaluate the performance of DREAM in reconstructing noisy PET data, we reconstructed 840 PET projection data sets with added Poisson noise. The final reconstructed images and their corresponding objective metrics were obtained and compared with the results from other methods.

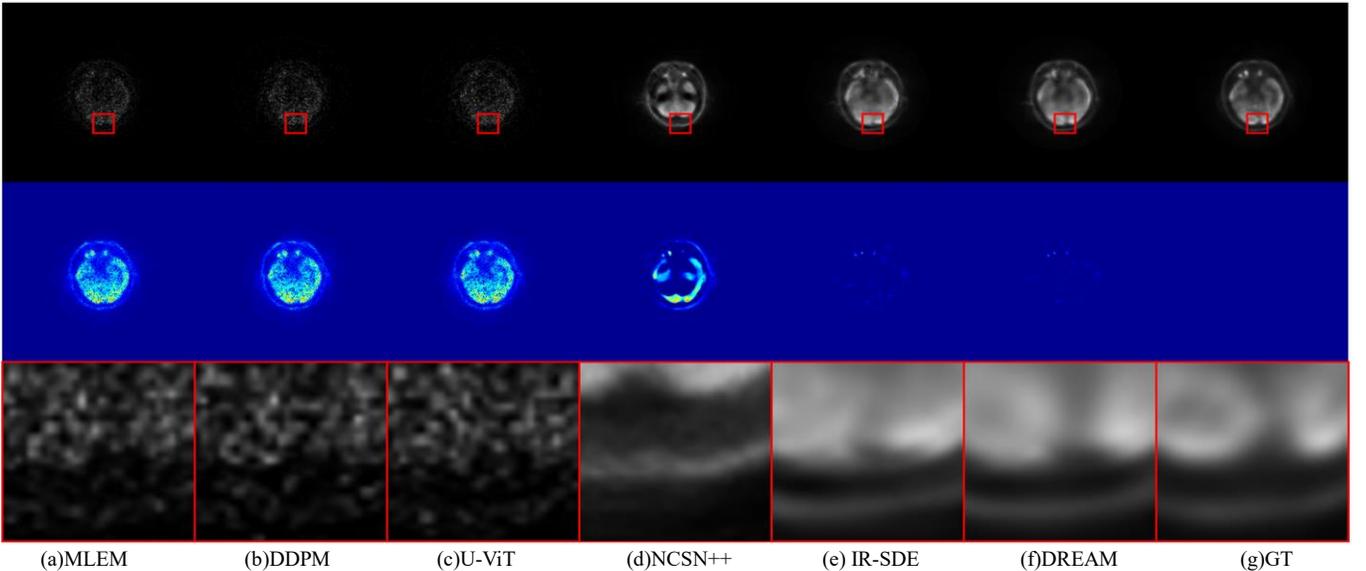

(a) MLEM    (b) DDPM    (c) U-ViT    (d) NCSN++    (e) IR-SDE    (f) DREAM    (g) GT

Fig.4. Reconstruction results for PET images using different methods. (a)-(f) show the reconstruction results and residual maps for various comparison methods and DREAM. (g) presents the noise-free reconstruction. The second row depicts the residuals between the reference and reconstructed images.

Table I presents the quantitative results of DREAM and various comparison methods. As shown, the reconstruction results exhibit two extremes. Compared to the MLEM method, NCSN++ shows some improvement, but its results are relatively limited. In contrast, DDPM and U-ViT perform poorly, with even a regression in performance. In comparison, DREAM achieves the best results, with PSNR and SSIM improving by 13.45dB and 0.1470, respectively, over MLEM, representing a significant enhancement. Furthermore, although IR-SDE performs well, our method still leads with a PSNR advantage of 1.35dB.

The results from the images and residual maps are consistent with the analysis above. Both IR-SDE and DREAM exhibit good reconstruction performance, with DREAM slightly outperforming IR-SDE in terms of smaller errors. In contrast, DDPM and U-ViT fail to effectively surpass MLEM. For DDPM and U-ViT, the dense points along the edges indicate that the reconstructed PET images are not as good as those from MLEM. Both DDPM and U-ViT are unsupervised models, which require large datasets to effectively learn image priors. Due to the relatively small size of our dataset, these models struggle to capture the necessary priors, resulting in poorer reconstruction quality. NCSN++ demonstrates some reconstruction ability in the center of the image, but its performance deteriorates significantly in the surrounding regions. DDPM and U-ViT show almost no reconstruction capability; both the images and residual maps are similar to MLEM. The explanation for this phenomenon is not difficult to understand: both DDPM and U-ViT are unsupervised deep learning methods, which face significant challenges in PET image reconstruction tasks.

In contrast, the diffusion model in our approach learns the data's latent space, a representation of the data in a compressed, lower-dimensional form that captures its essential structures. Specifically, we use SMCP within this latent space. SMCP refers to a technique where, during the reconstruction phase, the model iteratively predicts and refines the latent space representation of the image, ensuring consistency and improving accuracy. The reconstruction phase uses high-quality latent space data to assist the transformer in generating the final image, resulting in better reconstruction quality and faster reconstruction speed.

Additionally, analysis of the zoomed-in rectangular region, marked with a red box, provides further insights into the performance of each method. In this extracted region, DREAM preserves the finer details and edges, closely matching the ground truth (GT). The residual map shows minimal error, indicating high reconstruction accuracy. IR-SDE shows some structure but introduces blurring and inconsistencies, particularly around the edges, as shown by higher error values in the residual map. DDPM and U-ViT exhibit significant distortions, with noticeable noise and artifacts in both the images and the residual maps. NCSN++, while showing some reconstruction ability in the center of the region, deteriorates in the surrounding areas, with visible blurring and higher error values. These findings further confirm that DREAM outperforms DDPM, U-ViT, and NCSN++ in preserving fine details and minimizing artifacts, especially in critical areas of the image.

The subjective evaluation and objective quantification of the final reconstructed images above demonstrate the excellent PET reconstruction capability of DREAM. To further showcase the denoising reconstruction performance of DREAM, we present the sinogram data generated by each comparison method in Table II and Fig. 5.

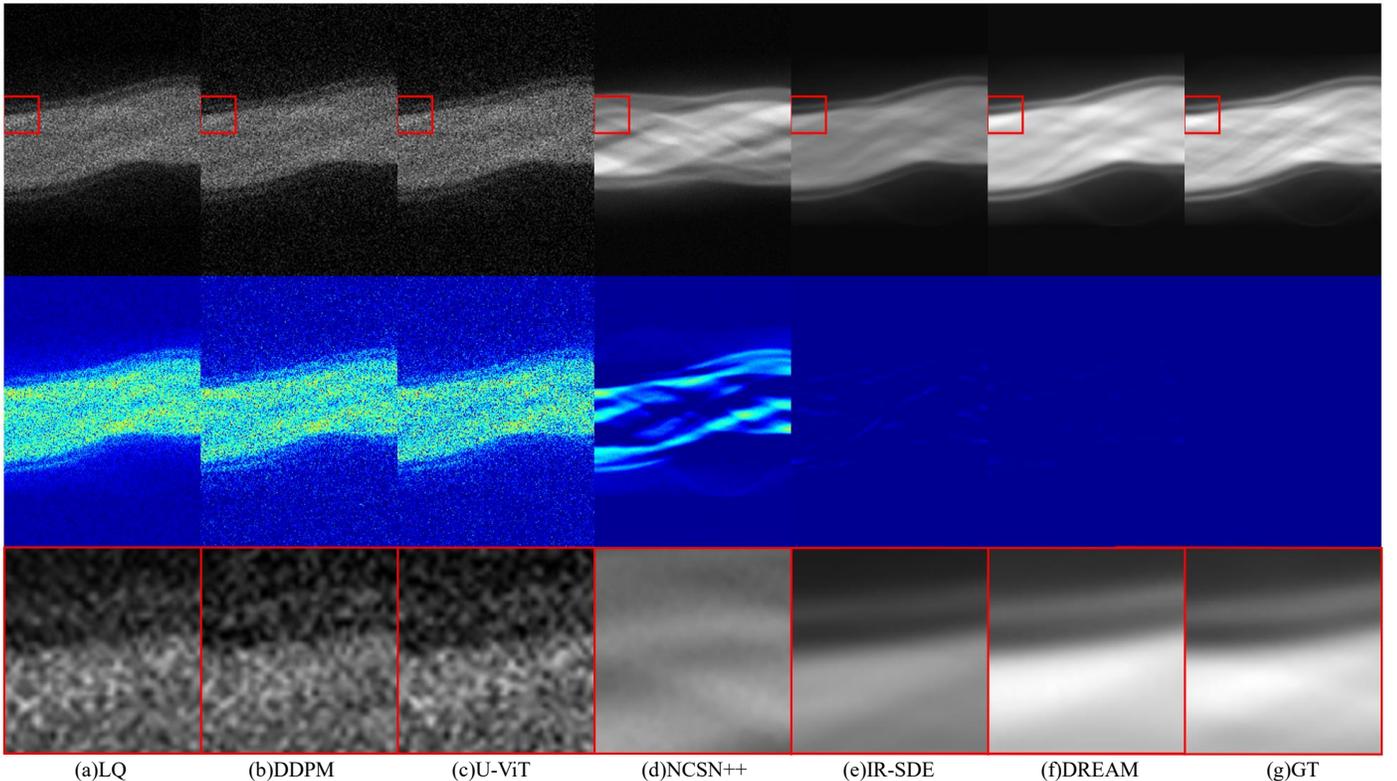

(a)LQ    (b)DDPM    (c)U-ViT    (d)NCSN++    (e)IR-SDE    (f)DREAM    (g)GT

**Fig.5.** Reconstruction results for PET sinograms using different methods. (a)-(f) show the sinogram data generated by various comparison methods and DREAM. (g) presents the noise-free sinogram. The second row depicts the residuals between the reference and reconstructed images.

Table II presents the quantitative results of DREAM and various comparison methods for sinogram data reconstruction. As shown, the reconstruction results exhibit significant differences. Compared to the other methods, DREAM achieves the best results, with PSNR and SSIM improving by 11.59 dB and 0.035, respectively, over IR-SDE, representing a substantial enhancement. Furthermore, the MSE for DREAM is significantly lower than the other methods, indicating better reconstruction accuracy.

These quantitative results are further supported by visual inspection of the sinogram and residual maps. Clearly, the images from DREAM are much more similar to the GT, and the residual maps confirm this, with only minimal errors. The images from IR-SDE show good overall contours and details, but the visual appearance is relatively dull, and the residual map also reveals some errors. For DDPM and U-ViT, the top and bottom sections of the images show significant noise compared to low quality (LQ), with corresponding areas in the residual maps showing increased brightness. Both methods perform poorly in denoising PET Sinogram data.

TABLE II
RECONSTRUCTION PSNR/SSIM/MSE FROM SINOGRAM DATA USING DIFFERENT METHODS.

| Methods | DDPM | U-ViT | NCSN++ | IR-SDE | **DREAM** |
|---|---|---|---|---|---|
| PSNR | 13.69 | 13.86 | 18.46 | 23.85 | **35.44** |
| SSIM | 0.115 | 0.119 | 0.527 | 0.931 | **0.966** |
| MSE | 0.0444 | 0.0427 | 0.0147 | 0.0064 | **0.0003** |

In particular, for DDPM and U-ViT, the dense dots along the edges of the sinogram indicate poor reconstruction quality. This phenomenon may be attributed to the fact that DDPM and U-ViT are unsupervised deep learning models. DDPM is a diffusion model, while U-ViT is transformer-based. Unsupervised learning methods usually require large datasets to achieve optimal performance, as they lack paired data to learn prior information. Given the relatively small size of our dataset, these unsupervised models struggle to effectively capture the underlying image priors, resulting in suboptimal reconstruction outcomes.

NCSN++, while demonstrating some reconstruction ability in the center of the image, suffers from significant blurring and deterioration in the surrounding areas, as seen in the residual map with larger errors. The performance of NCSN++ in preserving finer details and edges is inferior to that of DREAM. Overall, DREAM outperforms most methods with a significant margin, even surpassing IR-SDE, delivering superior results. DREAM is a reconstruction method with strong capabilities in noise suppression and artifact reduction.

In addition, the analysis of the extracted rectangular region, zoomed-in and highlighted with a red box, further highlights the performance of each method. As shown in the images of the rectangular region, DREAM consistently maintains sharp contours and details, closely matching the GT. In contrast, the images from IR-SDE exhibit some blurring and slight misalignment with the true structure. DDPM and U-ViT exhibit much more noticeable distortions in this region, with visible noise and artifacts that are not present in the GT or in DREAM. NCSN++ also shows some blurring and misalignment in this region, particularly around the edges. This further demonstrates the effectiveness of our approach in handling fine details and preserving the integrity of the image during reconstruction.

### E. Profile Lines Analysis

To evaluate the edge-preserving capability of different methods, we compared the contour lines of the reconstruction results from various methods. In Fig. 6, a contour line passing through the patient's brain region is plotted, visually illustrating the degree of alignment with the GT. MLEM, NCSN++, DDPM, and U-ViT fail to follow the GT trend in most regions, especially in the central valley area, where their performance is poor. However, MLEM and NCSN++ show some recovery near the peak to the right of the point at 100. In contrast, both IR-SDE and Ours closely follow the GT contour line, demonstrating good edge-preserving capability. However, in the region to the left of 100, IR-SDE fails to recover the contour, while DREAM closely matches the GT. Overall, DREAM demonstrates the most accurate contour generation for PET images.

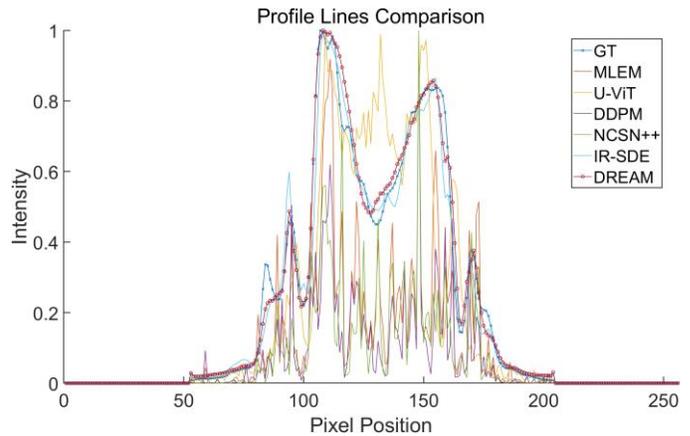

**Fig. 6.** Comparison of PET image profiles between the reference and other four methods on PET image reconstruction data.

Similarly, we need to evaluate the sinogram data generated by different methods. In Figure 7, we plot the contour lines of the results generated by each method, passing vertically through the sinogram images. From the ends of the images, it is evident that the contour lines for LQ, NCSN++, DDPM, and U-ViT are disordered, indicating poor denoising performance from these methods, with even higher peak values compared to LQ. In the middle section, only the contour line of NCSN++ closely follows the GT trend, showing some reconstruction capability, while DDPM and U-ViT remain similar to LQ. In contrast, DREAM and IR-SDE exhibit stronger reconstruction abilities, with DREAM clearly outperforming IR-SDE, and its advantage over IR-SDE is even more pronounced compared to Figure 6. Specifically, at the first peak, the central valley, and the peak to the right of 150, IR-SDE fails to recover the contours, whereas Ours nearly matches the GT. Additionally, at the peak to the left of 150, each method has its own advantages. IR-SDE is closer to the GT curve but has lower values, confirming its lower sinogram image brightness. DREAM, on the other hand, has values closer to the GT and presents a smoother curve, although it loses the original shape of the GT. Overall, the sinogram data generated by DREAM has the most accurate contour.

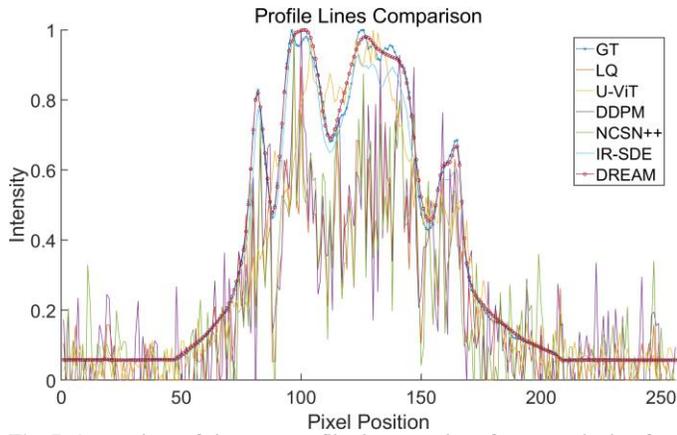

**Fig. 7.** Comparison of sinogram profiles between the reference and other four methods on sinogram reconstruction data.

### F. Ablation Study

Through the ablation study, a better understanding of the impact of each component on DREAM's performance, as well as their role within the entire model, can be attained. To further investigate the impact of certain factors on the performance of our reconstruction method, we primarily discuss two factors: the different mask addition methods, including random and hierarchical random masks, and the proportion of mask addition.

the random mask area was consistently set to 10%.

TABLE III
RECONSTRUCTION PSNR/SSIM/MSE USING DIFFERENT COMPONENTS.

| Methods | (w/o) All Masks | (w/o) Hierarchical Mask | **DREAM** |
|---|---|---|---|
| PSNR | 33.25 | 33.83 | **34.57** |
| SSIM | 0.952 | 0.953 | **0.955** |
| MSE | 0.0005 | 0.0004 | **0.0003** |

Table III and Fig.8 present the quantitative reconstruction results for each method, including PSNR, SSIM, and MSE. As shown, compared to the method without any masks, adding only the hierarchical mask results in a slight improvement in PSNR, with an increase of nearly 0.6 dB. However, the SSIM improvement is minimal. When both random and hierarchical masks are combined, the reconstruction quality improves significantly. For the dual-mask method, PSNR increases by 1.32 dB, and SSIM improves by 0.0027, demonstrating a marked enhancement in image quality compared to both the "no mask" and "hierarchical mask only" methods. Overall, the results indicate that combining random masks with a progressive hierarchical mask yields the best performance in terms of both PSNR and SSIM, while also minimizing MSE. This further supports the effectiveness of our approach in improving reconstruction quality by integrating both mask types.

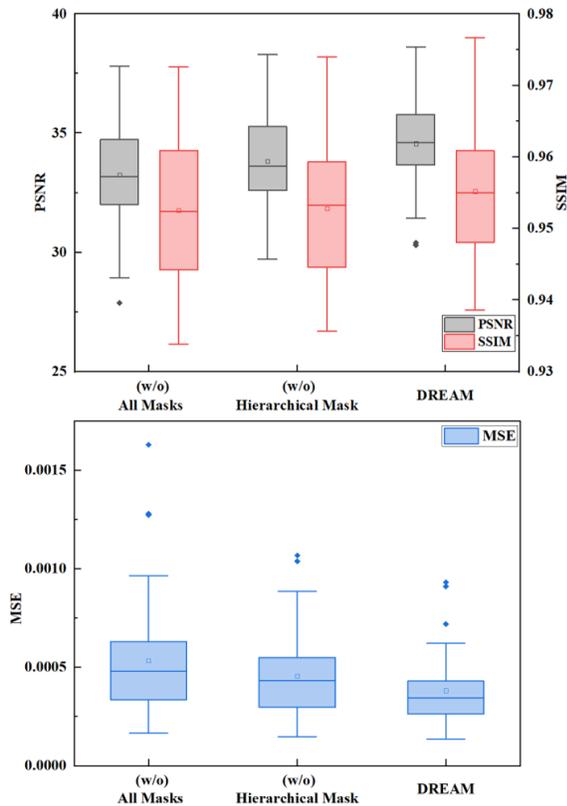

**Fig.8.** Boxplots of PSNR/SSIM/MSE among different mask algorithms.

As stated in the Motivation section of Chapter 3, random masks are a key contribution of our approach. To evaluate the impact of different mask addition strategies, we trained three models for the experiment: one without random masks, one with only hierarchical masks, and one with random masks combined with progressive hierarchical masks. In all experiments,

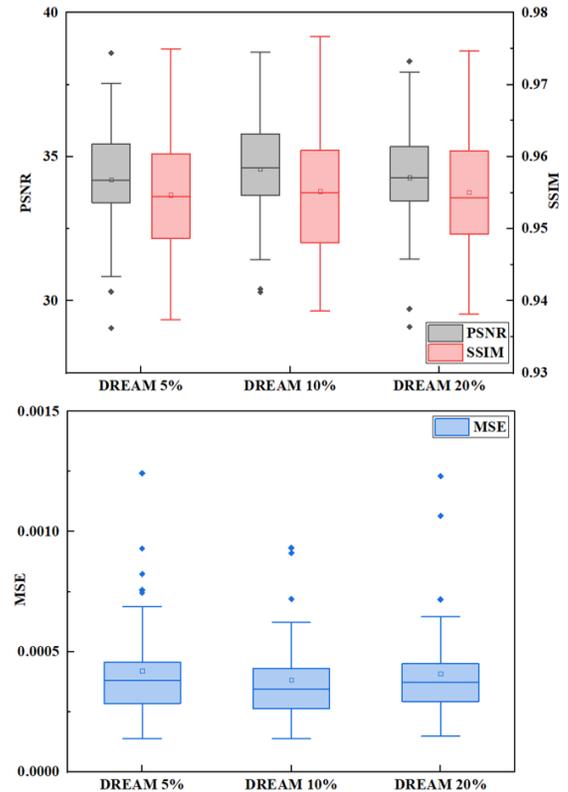

**Fig.9.** Boxplots of PSNR/SSIM/MSE among different mask levels.

Additionally, we investigated the impact of the random mask ratio on image reconstruction results. As shown in Table IV and Fig.9, the best reconstruction performance is achieved with a 10% random mask ratio, with PSNR, SSIM, and MSE all showing optimal values at this ratio. The PSNR is 34.57, SSIM is 0.955, and MSE is 0.0003, indicating high-quality reconstruction. However, when the ratio is decreased to 5% or increased to 20%, the reconstruction quality deteriorates, with a noticeable decrease in PSNR and MSE, indicating a reduction in image quality. Specifically, at 5%, the PSNR drops to 34.20, and at 20%, it returns to 34.28, which is lower than the result for 10%. These findings confirm that the reconstruction quality does not improve with an increase in the random mask ratio, but instead, the optimal results are obtained around a 10% ratio.

TABLE IV
RECONSTRUCTION PSNR/SSIM/MSE FOR DIFFERENT MASK LEVEL.

| Mask Level | 5% | **10%** | 20% |
|---|---|---|---|
| PSNR | 34.20 | **34.57** | 34.28 |
| SSIM | 0.954 | **0.955** | 0.955 |
| MSE | 0.0004 | **0.0003** | 0.0004 |

In all the experiments above, $N$ was always set to the optimal value of 2. To further validate the impact of channel number under high-dimensional stacking, we tested different values of $N$, specifically $N = 2, 3, 4$. In each case, one channel did not include random masks, while the remaining $(N-1)$ channels contained masked sinogram data. As shown in Table V, the best reconstruction performance was achieved when $N = 3$. This improvement is likely due to the optimal balance between the number of masked and unmasked channels, allowing the model to better capture both local details and broader spatial relationships.

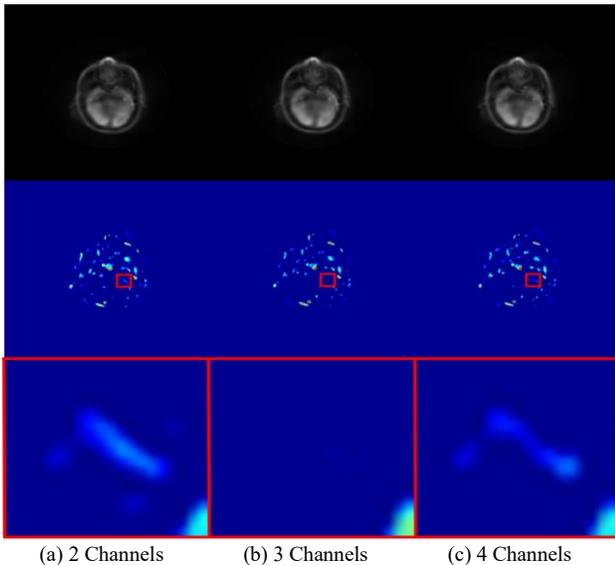

(a) 2 Channels    (b) 3 Channels    (c) 4 Channels

**Fig.10.** Reconstruction results for PET images using different channel numbers. The second row depicts the residuals between the reference and reconstructed images.

TABLE V
RECONSTRUCTION PSNR/SSIM/MSE FOR DIFFERENT CHANNEL NUMBERS.

| Channel Number | 2 | **3** | 4 |
|---|---|---|---|
| PSNR | 34.32 | **34.57** | 34.07 |
| SSIM | **0.955** | 0.955 | 0.954 |
| MSE | 0.0004 | **0.0003** | 0.0004 |

In summary, we investigated the effects of different masking strategies, the amount of mask applied, and the number of channels on the marginal effects of high-dimensional stacking. Through these analyses, the DREAM model reached its optimal performance, demonstrating the critical role of these factors in achieving high-quality PET image reconstruction.

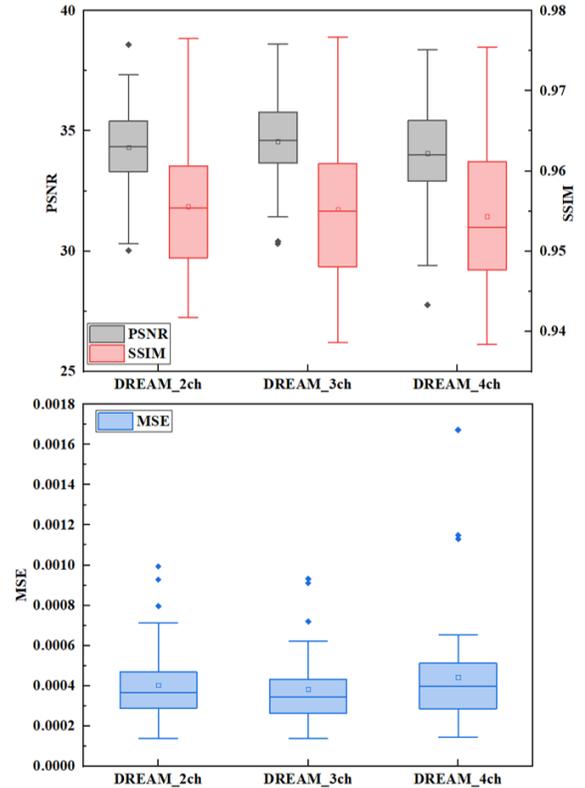

**Fig.11.** Boxplots of PSNR/SSIM/MSE among different channel numbers.

## V. DISCUSSION AND CONCLUSIONS

This study introduces DREAM, a novel PET image reconstruction framework that integrates advanced mask mechanisms within the diffusion transformer model. By embedding masks into both the sinogram domain and latent space, DREAM transforms traditional 2D processing into a 3D paradigm through high-dimensional stacking, enriching the solution space and enabling the model to capture intricate spatial relationships. The mask-driven latent space, built on sinogram-driven and mask-driven compact priors, accelerates the diffusion process by focusing on compact, high-priority regions, reducing computational costs while preserving essential data characteristics. Additionally, a hierarchical masking strategy enables progressive learning, transitioning from localized feature extraction to global pattern recognition, achieving a balance between detail retention and structural comprehension. Future research should aim to investigate the impact of mask shapes on PET reconstruction quality, as this study primarily focuses on square masks. Understanding how varying mask geometries influence reconstruction performance could provide valuable insights into optimizing mask-based methodologies. Additionally, future work could explore the application of masks in other medical imaging domains, such as PET tracer separation, to further enhance their versatility and efficacy in addressing diverse challenges in medical imaging.

# REFERENCES

[1] M. E. Phelps, "Positron emission tomography provides molecular imaging of biological processes," *PNAS.* vol. 97, no. 16, pp. 9226-9233, 2000.
[2] R. S. Cherry, "Fundamentals of positron emission tomography and applications in preclinical drug development," *J. Clin. Pharmacol*, vol. 41, no. 5, pp. 482-491. 2001.
[3] A. M. Alessio, P. E. Kinahan, M. Vivek, G. Victor, A. Lisa, and M. T. Parisi, "Weight-based, low-dose pediatric whole-body PET/CT protocols," *J. Nucl. Med*, vol. 50, no. 10, pp. 1570-1578, 2009.
[4] W. D. Townsend, "Multimodality imaging of structure and function," *Phys. Med. Biol*, vol. 53, no. 4, pp. R1, 2008.
[5] J. D. Brenner, and E. J. Hall, "Computed tomography—an increasing source of radiation exposure," *New Eng. J. Med*, vol. 357, no. 22, pp. 2277-2284, 2007.
[6] H. Kirpalani, and N. Claude, "Radiation risk to children from computed tomography," *Pediatrics*, vol. 121, no. 2, pp. 449-450, 2008.
[7] M. S. Pearce, J. A. Salotti, and M. P. Little, "Radiation exposure from CT scans in childhood and subsequent risk of leukaemia and brain tumours: a retrospective cohort study," *Pediatr. Radiol.*, vol. 43, no. 5, pp. 517-518, 2013.
[8] S. Mattsson, and S. Marcus, "Radiation dose management in CT, SPECT/CT and PET/CT techniques," *Radiat. Prot. Dosim.*, vol. 147, no. 1-2, pp. 13-21, 2011.
[9] R. Boellaard, M. J. O'Doherty, W. A. Weber, F. M. Mottaghy, M. N. Lonsdale, S. G. Stroobants, W. J. G. Oyen, *et al.,* "FDG PET and PET/CT: EANM procedure guidelines for tumour PET imaging: version 1.0," *Eur. J. Nucl. Med. Mol. Imaging*, vol. 37, no. 2, pp. 181-200, 2010.
[10] B. G. Wang, and Y. J. Qi, "PET image reconstruction using kernel method," *IEEE Trans. Med. Imaging*, vol. 34, no. 1, pp. 61-71, 2014.
[11] R. D. Schaart, "Physics and technology of time-of-flight PET detectors," *Phys. Med. Biol.*, vol. 66, no. 9, 2021.
[12] J. M. Geusebroek, A. W. M Smeulders, and J. Van De Weijer. "Fast anisotropic gauss filtering," *IEEE Trans Image Process*, vol. 12, no. 08, pp. 938-943, 2003.
[13] C. Wang, Z. Hu, P. Shi, and H. Liu, "Low dose PET reconstruction with total variation regularization," *EMBC*, pp. 1917-1920, 2014.
[14] X. Yu, *et al.* "Low dose PET image reconstruction with total variation using alternating direction method" *PloS One*, pp.0166871, 2016.
[15] J. Dutta, Q. Li and R. M. Leahy, "Non-local means denoising of dynamic PET images," *PloS One*, vol. 8, no. 3, pp. 1-15, 2013.
[16] H. Arabi, and H. Zaidi. "Spatially guided nonlocal mean approach for denoising of PET images." *Med. Phys.*, vol. 47, no. 4, pp. 1656-1669, 2022.
[17] C. L. Yann, B. Yoshua and H. Geoffrey, "Deep learning," *NATUAS*, vol. 521, no. 7553, pp. 436-444, 2015.
[18] I. Goodfellow, Bengio, Y. Bengio and A. Courville, "Deep learning," *MIT Press.*, 2016.
[19] G. Kuang, J. H. Guan, K. Kim, X. Z. Zhang, J. Yang, Y. Seo, G. E. Fakhri, J. Y. Qi, and Q. Z. Li, "Iterative PET image reconstruction using convolutional neural network representation," *IEEE Trans. Med. Imaging*, vol. 8, no. 3, pp. 675-685, 2018.
[20] J. Xu, E. Gong, J. Pauly and G. Zaharchuk, "200x low-dose PET reconstruction using deep learning," *arXiv preprint arXiv*:1712. 04119, 2017.
[21] Z. Y. Peng, F. W. Zhang, J. Z. Sun, Y. Du, Y. Wang, and G. S. P. Mok, "Preliminary deep learning-based low dose whole body PET denoising incorporating CT information," *In IEEE NSS/MIC*, pp. 1-2, 2022.
[22] K. Gong, J. H. Guan, C. Liu, and J. Y. Qi, "PET image denoising using a deep neural network through fine tuning," *IEEE Trans. Radiat. Plasma Med. Sci*, vol. 3, no. 2, pp. 153-161, 2018.
[23] L. Zhou, J. D. Schaefferkoetter, I. W. Tham, G. Huang and J. Yan, "Supervised learning with cyclegan for low-dose FDG PET image denoising," *Med. Image Anal.,* vol. 65, no. 1, p. 101770, 2020.
[24] S. Pan, E. Abouei, J. Peng, J. Qian, J. F. Wynne, T. Wang, C. Chang, *et al.,* "Full-dose PET synthesis from low-dose PET using 2D high efficiency denoising diffusion probabilistic model," *In Medical Imaging 2024: Clin. Biomed. Imaging*, vol. 12930, pp. 428-435, 2024.
[25] S. Iizuka, E. Simo-Serra, and M. Ishii, "Globally and locally consistent image completion," *ACM Trans. Graph.*, vol. 36, no. 4, p. 107, 2017.
[26] D. Pathak, P. Krahenbuhl, J. Donahue, T. Darrell, and A. A. Efros, "Context encoders: feature learning by inpainting," *Proc. IEEE Conf. Comput. Vis. Pattern Recognit.*, pp. 2536-2544, 2016.
[27] H. Xiang, Q. Zou, M. A. Nawaz, X. Huang, F. Zhang, and H. Yu, "Deep learning for image inpainting: a survey," *Pattern Recognit.*, vol. 134, p. 109046, 2023.
[28] J. Long, E. Shelhamer, and T. Darrell, "Fully convolutional networks for semantic segmentation," *Proc. IEEE Conf. Comput. Vis. Pattern Recognit.*, pp. 3431-3440, 2015.
[29] R. Wang, T. Lei, R. Cui, B. Zhang, H. Meng, and A. K. Nandi, "Medical image segmentation using deep learning: a survey," *IET Image Process.*, vol. 16, no. 5, pp. 1243-1267, 2022.
[30] K. Gong, K. Johnson, G. E. Fakhri, Q. Li, and T. Pan, "PET image denoising based on denoising diffusion probabilistic model," *Eur. J. Nucl. Med. Mol. Imaging*, vol. 51, no. 2, pp. 358-368, 2024.
[31] C. Jiang, Y. Pan, M. Liu, L. Ma, X. Zhang, J. Liu, X. Xiong, and D. Shen, "PET-diffusion: unsupervised PET image enhancement based on the latent diffusion model," *In MICCAI*, vol. 14220, pp. 3-12, 2023.
[32] Y. Luo, Y. Wang, C. Zu, B. Zhan, X. Wu, J. Zhou, D. Shen, and L. Zhou, "3D transformer-GAN for high-quality PET reconstruction," *In MICCAI*, vol. 12906, pp. 276-285, 2021.
[33] R. Hu and H. Liu, "TransEM: Residual Swin-transformer based regularized PET image reconstruction," *In MICCAI,* vol. 13434, pp. 184-193, 2022.
[34] A. Dosovitskiy, L. Beyer, A. Kolesnikov, D. Weissenborn, X. Zhai, T. Unterthiner, M. Dehghani, M. Minderer, G. Heigold, S. Gelly, *et al.,* "An image is worth 16x16 words: transformers for image recognition at scale," *In ICLR*, 2021.
[35] S. Khan, M. Naseer, M. Hayat, S. W. Zamir, F. S. Khan, and M. Shah, "Transformers in vision: a survey," *In CSUR,* vol. 54, no. 10, pp. 1-41, 2022.
[36] J. Cui, P. Zeng, X. Zeng, P. Wang, X. Wu, J. Zhou, *et al*., "TriDoFormer: a triple-domain transformer for direct PET reconstruction from low-dose sinograms," *In Proc. Int. Conf. Med. Image Comput. Comput.-Assist. Interv.*, Cham, Switzerland: Springer, pp. 184-194, 2023.
[37] B. Huang, X. Liu, L. Fang, Q. Liu, and B. Li, "DTM: diffusion transformer model with compact prior for low-dose PET reconstruction," *arXiv preprint arXiv*:2303.06155, 2023.
[38] T. Yu, Z. Zhang, C. Lan, Y. Lu, and Z. Chen, "Mask-based latent reconstruction for reinforcement learning," *NeurIPS,* 2022.
[39] H. Wang, Y. Tang, Y. Wang, J. Guo, Z. H. Deng, and K. Han, "Masked image modeling with local multi-scale reconstruction," *arXiv preprint arXiv*:2303.06538, 2023.
[40] G. Liu, F. A. Reda, K. J. Shih, T. C. Wang, A. Tao, and B. Catanzaro, "Image inpainting for irregular holes using partial convolutions," *In ECCV*, pp. 85-100, 2018.
[41] O. Ronneberger, P. Fischer, and T. Brox, "U-Net: convolutional networks for biomedical image segmentation," *In MICCAI*, pp. 234-241, 2015.
[42] J. Long, E. Shelhamer, and T. Darrell, "Fully convolutional networks for semantic segmentation," *Proc. IEEE Conf. Comput. Vis. Pattern Recognit.* pp. 3431-3440, 2015.
[43] B. Huang, X. Liu, L. Fang, Q. Liu, and B. Li, "Diffusion transformer model with compact prior for low-dose PET reconstruction," *arXiv preprint arXiv*:2407.00944, 2024.
[44] D. Hendrycks and K. Gimpel, "Gaussian error linear units (GELUs)," *arXiv preprint arXiv*:1606. 08415, 2016.
[45] H. M. Hudson and R. S. Larkin, "Accelerated image reconstruction using ordered subsets of projection data," *IEEE Trans. Med. Imaging*, vol. 13, no. 4, pp. 601-609, 1994.
[46] F. Bao, S. Nie, K. Xue, Y. Cao, C. Li, H. Su, and J. Zhu, "All are worth words: A ViT backbone for diffusion models," *Proc. IEEE/CVF Conf. Comput. Vis. Pattern Recognit.*, 2023.
[47] Y. Song, J. Sohl-Dickstein, D. P. Kingma, A. Kumar, S. Ermon, and B. Poole, "Score-based generative modeling through stochastic differential equations," *In ICLR*, 2021.
[48] Z. Luo, F. K. Gustafsson, Z. Zhao, J. Sjölund, and T. B. Schön, "Image restoration with mean-reverting stochastic differential equations," *Proc. 40th Int. Conf. Mach. Learn*, vol. 202, no. 957, pp. 23045-23066, 2023.